\documentclass[lettersize,journal]{IEEEtran}

\usepackage{amsmath}
\usepackage{amsfonts}
\usepackage{bm}
\usepackage{mathptmx}
\usepackage[T1]{fontenc}
\usepackage[mathscr]{eucal}

\usepackage[table]{xcolor}
\definecolor{tabred}{HTML}{d62728}
\definecolor{darkgreen}{HTML}{006400}

\usepackage{graphicx}
\usepackage{graphics}
\usepackage{epsfig}
\usepackage{epstopdf}
\usepackage{float}
\usepackage{subfig}
\usepackage{overpic}
\usepackage{adjustbox}

\usepackage{booktabs}
\usepackage{multirow}
\usepackage{threeparttable}
\usepackage{tabularx}
\usepackage{array}
\usepackage{makecell}

\usepackage[ruled,vlined]{algorithm2e}

\usepackage{tikz}
\usepackage{pgfplots}
\pgfplotsset{compat=1.15}

\usepackage{siunitx}

\usepackage{textcomp}
\usepackage{stfloats}
\usepackage{verbatim}
\usepackage{autobreak}
\usepackage{balance}
\usepackage{enumerate}

\usepackage{cite}

\usepackage{url}
\usepackage{xurl}
\usepackage{hyperref}
\hypersetup{
    colorlinks=true,
    breaklinks=true,         
    linkcolor=[rgb]{1.0, 0.4, 0.4}, 
    urlcolor=magenta,               
}
\begin{document}

\title{Optimal-Horizon Social Robot Navigation in Heterogeneous Crowds}
\author{Jiamin Shi$^{1}$, Haolin Zhang$^{1}$, Chongfei Chen, Yuchen Yan$^{1}$, Shitao Chen$^{\dag,}$$^{1}$, Jingmin Xin$^{1}$, Nanning Zheng$^{1}$ 
\thanks{$^{1}$J. Shi, H. Zhang, Y. Yan, S. Chen, J. Xin and N. Zheng are with National Key Laboratory of Human-Machine Hybrid Augmented Intelligence, Nation Engineering Research Center for Visual Information and Applications, and Institute of Artificial Intelligence and Robotics, Xi'an Jiaotong University, Xi'an, Shaanxi 710049, P.R. China;
        {\tt\small 3120105180@stu.xjtu.edu.cn;
        chenshitao@xjtu.edu.cn;
        jxin, nnzheng@mail.xjtu.edu.cn}}
\thanks{$^{\dag}$S. Chen is the corresponding author.}%
}
\maketitle
\begin{abstract}
Navigating social robots in dense, dynamic crowds is challenging due to environmental uncertainty and complex human-robot interactions. 
While Model Predictive Control (MPC) offers strong real-time performance, its reliance on a fixed prediction horizon limits adaptability to changing environments and social dynamics. Furthermore, most MPC approaches treat pedestrians as homogeneous obstacles, ignoring social heterogeneity and cooperative or adversarial interactions, which often causes the Frozen Robot Problem in partially observable real-world environments. 
In this paper, we identify the planning horizon as a socially conditioned decision variable rather than a fixed design choice. Building on this insight, we propose an optimal-horizon social navigation framework that optimizes MPC foresight online according to inferred social context. A spatio-temporal Transformer infers pedestrian cooperation attributes from local trajectory observations, which serve as social priors for a reinforcement learning policy that optimally selects the prediction horizon under a task-driven objective. The resulting horizon-aware MPC incorporates socially conditioned safety constraints to balance navigation efficiency and interaction safety. Extensive simulations and real-world robot experiments demonstrate that optimal foresight selection is critical for robust social navigation in partially observable crowds. Compared to state-of-the-art baselines, the proposed approach achieves a 6.8\% improvement in success rate, reduces collisions by 50\%, and shortens navigation time by 19\%, with a low timeout rate of 0.8\%, validating the necessity of socially optimal planning horizons for efficient and safe robot navigation in crowded environments. Code and videos are available at Under Review.
\end{abstract}


\section{Introduction}
\IEEEPARstart{S}{ocial} robot navigation in densely populated and dynamic environments is a critical capability for applications such as service robotics, autonomous delivery, and assistive technologies~\cite{everett2018motion, chen2017socially}. Safe and efficient navigation among moving pedestrians requires not only prediction of complex and uncertain environmental dynamics but also understanding and respecting diverse social norms and interactions~\cite{kretzschmar2016socially, kong2025socially}. This dual challenge remains a significant hurdle for existing navigation frameworks, particularly in partially observable and socially heterogeneous environments.

\begin{figure}[t]
    \centering
    \vspace{0cm}
    \setlength{\abovecaptionskip}{0cm}
    \includegraphics[width=0.9\linewidth]{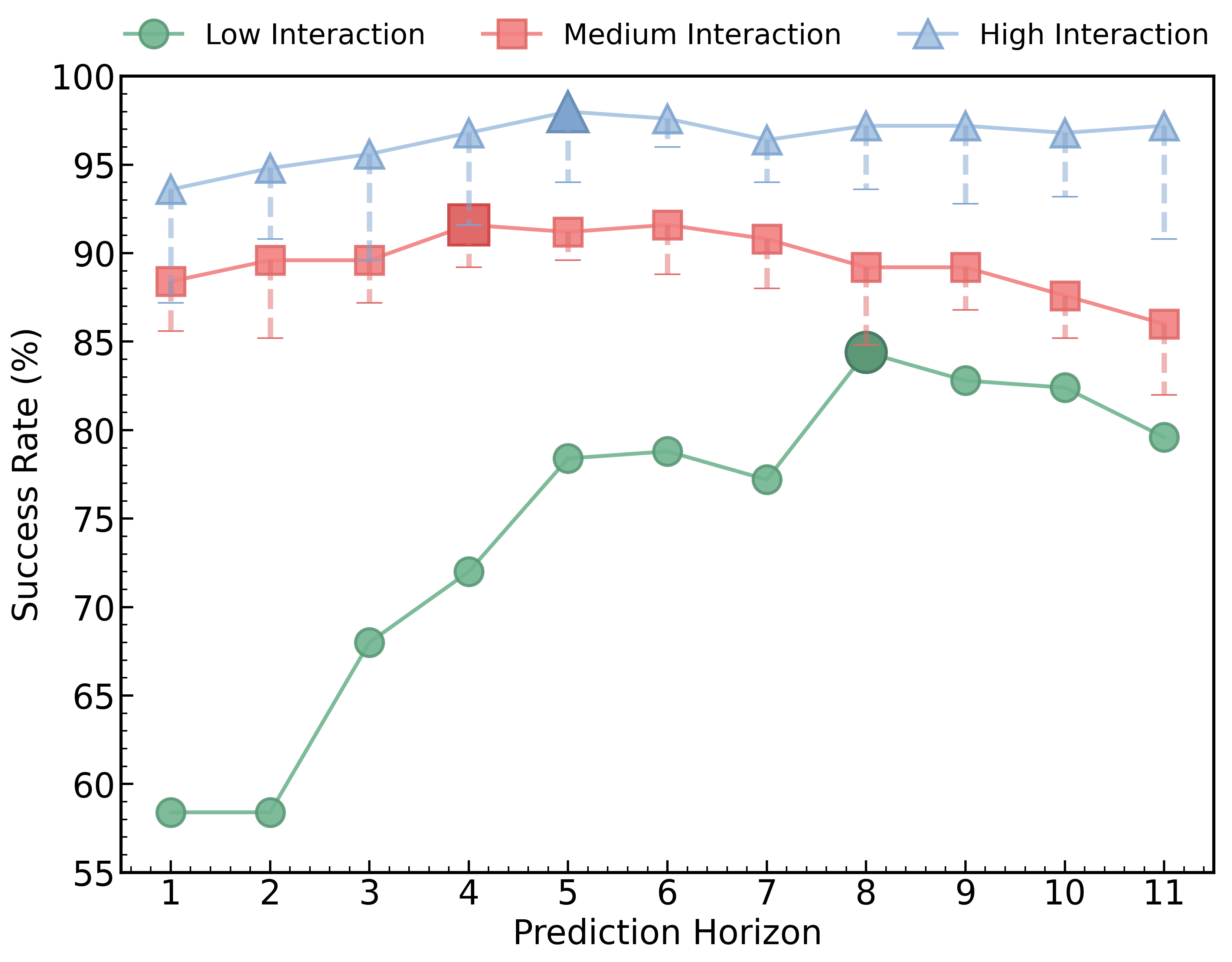}
    \caption{Navigation success rates under fixed MPC prediction horizons, averaged over 250 random trials. Solid lines indicate the average performance, while vertical error bars represent performance variations across different pedestrian behavior modes. Note that the low-interaction scenario lacks error bars as all pedestrians are consistently non-cooperative, resulting in a single behavior mode.}
    \label{fig:intro}
\end{figure}

Model Predictive Control (MPC) has been widely adopted in robot navigation due to its ability to generate dynamically feasible trajectories while respecting physical and safety constraints~\cite{samavi2025sicnav}. Its inherent real-time replanning and controllability make it well suited for dynamic scenarios involving moving agents~\cite{icra24_coop_mpc}. However, most MPC-based navigation methods rely on a fixed prediction horizon, limiting their flexibility to adapt to varying interaction densities and social complexities of the environment~\cite{icra24_drcc_mpc, shamsah2024socially}. This assumption implicitly treats future interaction complexity as stationary over time. While adaptive-horizon MPC has been studied in contexts such as autonomous racing primarily for computational efficiency~\cite{kabzan2019learning}, its potential role in social interaction reasoning remains largely unexplored.

In socially interactive and partially observable environments, a fixed planning horizon often leads to suboptimal behaviors. Short horizons tend to favor reactive and myopic decisions that fail to anticipate complex social interactions, whereas long horizons may induce overly conservative behaviors in the presence of uncertain human dynamics. Such rigidity can significantly degrade navigation performance and constitutes a key underlying cause of the well-known Frozen Robot Problem~\cite{FRP}, where robots become overly cautious and cease to move despite the existence of feasible navigation paths. These observations indicate that no single fixed horizon can adequately accommodate the non-stationary nature of human--robot interactions, and that the planning horizon should be treated as a socially conditioned decision variable, specifically in response to pedestrians' cooperation tendencies.

Navigation strategies further often treat pedestrians as homogeneous dynamic obstacles, relying primarily on trajectory prediction rather than reasoning about underlying social behaviors~\cite{shamsah2024socially}. This simplification overlooks critical social heterogeneity, such as differing pedestrian intents and cooperative or adversarial tendencies, which directly impact navigation safety and efficiency. Although some studies have introduced social cost functions or proxemics-based constraints to account for interpersonal comfort~\cite{chen2020relational}, explicit modeling of heterogeneous pedestrian social characteristics remains largely unexplored within MPC-based navigation frameworks.

To address these challenges, we propose \textbf{Optimal-Horizon Social Robot Navigation}, a socially compliant MPC framework that explicitly treats the planning horizon as the primary control decision in response to latent social complexity. The core idea is to couple social attribute inference with \emph{optimal foresight control}, where the MPC prediction horizon is treated as an \emph{optimization variable} and selected online based on the inferred social context. A pretrained spatio-temporal Transformer infers pedestrian cooperation attributes from local trajectory observations, which are embedded into a heterogeneous state representation and serve as social priors for a reinforcement learning policy that selects the prediction horizon. The resulting optimal-horizon MPC further incorporates socially conditioned safety constraints that modulate collision avoidance margins based on inferred pedestrian behaviors. By dynamically adjusting safety distances, the framework enables efficient interaction with cooperative pedestrians while maintaining conservative buffers around non-cooperative individuals, effectively balancing navigation efficiency and long-term social safety.

The main contributions of this work are summarized as follows:
\begin{itemize}
  \item We identify planning horizon selection as a socially conditioned decision variable in robot navigation and formalize planning horizon selection as an \emph{optimal decision mechanism} to resolve social deadlocks in heterogeneous crowds.
  \item We develop an optimal-horizon MPC framework that integrates social attribute inference and reinforcement learning to determine the prediction horizon online according to social context.
  \item We design socially conditioned safety constraints that dynamically modulate collision avoidance strategies based on inferred pedestrian social attributes.
  \item We validate the proposed approach through extensive simulations and real-world robot experiments, demonstrating improved safety, efficiency, and robustness over state-of-the-art baselines in crowded, partially observable environments.
\end{itemize}

\section{Related Work}

\subsection{MPC with Fixed Prediction Horizons}
Model Predictive Control (MPC) has been widely adopted in robot navigation due to its ability to generate dynamically feasible trajectories under Control Barrier Functions (CBFs)~\cite{mestres2024distributed, tan2022distributed}. Adaptive MPC approaches have been explored~\cite{soman2025learning}, but they mainly focus on tuning control objectives or constraints for general performance and do not explicitly address the non-stationary interaction complexity induced by heterogeneous human crowds. In practice, most MPC-based social navigation frameworks still rely on fixed prediction horizons~\cite{samavi2025sicnav, shamsah2024socially}, implicitly assuming stationary interaction difficulty, which limits adaptability to varying crowd densities and social behaviors. In contrast, our method treats horizon selection as a decision variable and learns a dynamic horizon policy via reinforcement learning to adapt planning foresight to social complexity online.

\subsection{MPC in Socially Compliant Robot Navigation}
Socially compliant navigation requires robots to account for both physical feasibility and social expectations during human--robot interaction~\cite{everett2018motion, samavi2025sicnav}. Recent advances focus on human behavior modeling and trajectory prediction~\cite{le2024social, liu2026height}, with transformer-based methods showing strong spatio-temporal representation capability. However, these approaches primarily rely on predicted trajectories and still model pedestrians as homogeneous dynamic obstacles, lacking explicit reasoning about social heterogeneity such as cooperative versus non-cooperative behaviors, and thus remain susceptible to the Frozen Robot Problem in partially observable environments. By contrast, our approach explicitly infers pedestrian social attributes using a pretrained Transformer and integrates them into a heterogeneous MPC state representation with socially conditioned safety constraints.

\subsection{Learning-Augmented MPC Approaches}
Learning-augmented MPC has been proposed to improve adaptability by incorporating learning-based components into classical control frameworks. Reinforcement learning has been used to tune MPC cost functions or constraint parameters in general control tasks~\cite{nguyen2024model, stefanini2024efficient, kabzan2019learning}, but adaptive selection of MPC prediction horizons remains largely unexplored, particularly in social navigation contexts. Cooperative strategies in heterogeneous multi-agent navigation have also been studied~\cite{zhang2024heterogeneous}, yet these methods focus on robot--robot coordination rather than human--robot interaction and do not consider latent social attributes or planning foresight. The proposed framework fills this gap by jointly modeling pedestrian social attributes and horizon-adaptive planning within a unified, socially aware MPC formulation.

\section{Method}
\begin{figure*}[t]
    \centering
    \includegraphics[scale=0.35]{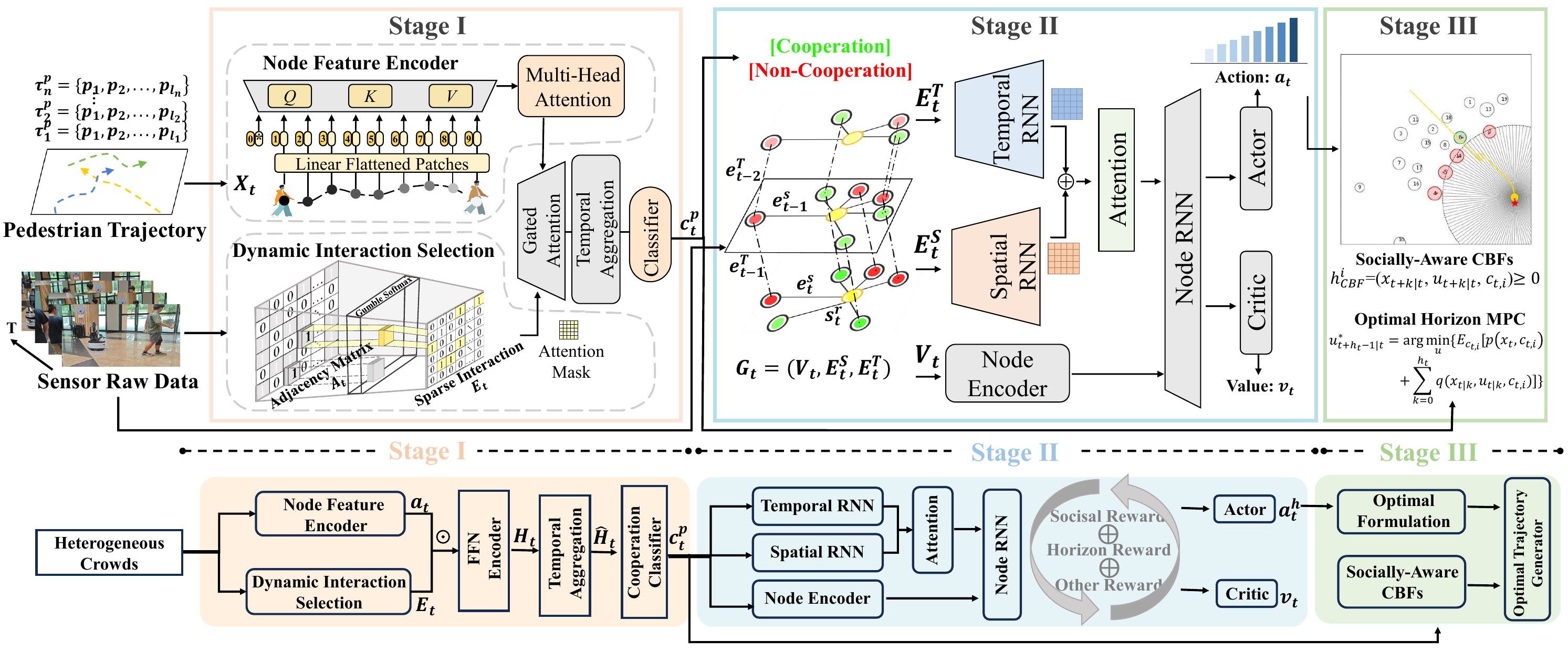}
    \caption{Overview of the Optimal-Horizon Social Robot Navigation framework. Stage I: The Spatio-Temporal Transformer infers pedestrian cooperation attributes $c_t^p$ from trajectory history. Stage II: The RL policy determines the optimal prediction horizon $h_t$ based on the inferred social graph $G_t$. Stage III: The DTCBF-constrained MPC generates safe, socially-compliant control commands $u_t^*$ by adapting its look-ahead depth and safety margins.}
    \label{fig:Framework}
\end{figure*}
\subsection{Overall Framework}
To address social navigation challenges in dense and dynamic environments, this paper proposes the Optimal-Horizon Social Robot Navigation framework, illustrated in Fig.~\ref{fig:Framework}. The framework explicitly models heterogeneous social interactions, enabling the robot to collaborate with cooperative pedestrians while cautiously avoiding non-cooperative ones, mitigating the Frozen Robot Problem under partial observability. It comprises three interrelated components: Spatio-Temporal Transformer for Pedestrian Cooperation Prediction, Reinforcement Learning for Dynamic Horizon Decision, and Optimal-Horizon MPC with Social Awareness.

\textbf{\textcolor{black}{Stage I:}} A spatio-temporal transformer takes pedestrian trajectories as input, constructs a multirelational spatio-temporal structure that captures both direct spatial proximity and indirect social influence patterns, and infers cooperation attributes $c^p_t$. The predicted attributes are embedded into individual pedestrian states, enabling the robot to distinguish between cooperative and non-cooperative pedestrians for socially aware navigation.

\textbf{\textcolor{black}{Stage II:}} The spatio-temporal graph $G_t = (V_t, E_t^S, E_t^T)$ is fed into a RL policy $\pi_\theta$ that dynamically determines the socially conditioned MPC prediction horizon $h_t$. Leveraging spatial and social information, the policy enables context-sensitive horizon decisions that adapt to crowd density and social dynamics in real time, improving planning flexibility and efficiency.

\textbf{\textcolor{black}{Stage III:}} The MPC receives the predicted cooperation attributes $c^p_t$ and the optimal prediction horizon $h_t$ to generate collision-free trajectories $\tau_t$. 
Socially-Aware CBFs adapt safety distances based on the inferred cooperation attributes $c^p_t$, while ensuring safe execution in heterogeneous social contexts.
 
To deploy in real-world scenarios, full observability is not assumed. Only pedestrians within the robot's sensor range are considered, and their positions and velocities are transformed into the robot coordinate system, with occlusions further constraining observations.

\subsection{Spatio-Temporal Transformer for Pedestrian Cooperation Prediction}
\label{sec:ST-Transformer}
This stage introduces an interaction-aware transformer to capture spatial interactions and temporal dynamics of pedestrian trajectories. The architecture comprises a multi-layer transformer encoder and a lightweight cooperation classifier.

Unlike standard transformers, this model employs a dynamic interaction selection mechanism to determine relevant pedestrian relationships. At time $t$, the $i$-th pedestrian trajectory $\tau_{i}^p=\{p_1,\dots,p_L\}$ is a sequence of past $L$ positions where $p \in \mathbb{R}^2$. These trajectories are embedded into features $x_{t,i}$; stacking $M$ visible pedestrians yields $X_t\in\mathbb{R}^{M\times d}$, where $d$ is the feature dimension. Interaction representation $I_t = (X_t, A_t)$ utilizes an adjacency matrix $A_t$ derived from raw sensor data.

The learnable selector computes interaction probabilities from $A_t$ and samples connections via the Gumbel-Softmax estimator \cite{jang2016categorical}:
\begin{equation}
\small
E_t = \text{Gumbel-Softmax}(\log(A_t + \epsilon), \mathcal{T}_c)
\label{eq:gumbel_selection}
\end{equation}
where $\mathcal{T}_c$ is the temperature parameter, $\epsilon$ is a stability constant, and $E_t$ represents the resulting sparse interaction edge matrix.

The attention mechanism operates on this structure $E_t$:
\begin{equation}
\small
\text{InterAttention}(Q,K,V,E_t)=\text{softmax}\!\Big(\frac{QK^\top}{\sqrt{d_k}}\Big) \odot E_t \cdot V
\label{eq:inter_attention}
\end{equation}
\begin{equation}
\small
H_t=\text{FFN}\!\big(X_t+\text{InterAttention}(X_tW_Q,X_tW_K,X_tW_V,E_t)\big)
\label{eq:transformer_output}
\end{equation}
where $\odot$ denotes element-wise multiplication, $d_k$ is the scaling dimension, and $W_Q, W_K, W_V$ are weight matrices. 

Temporal aggregation yields a pedestrian descriptor $\bar{H}_i = \frac{1}{L}\sum_{\ell=1}^{L} H_{\ell,i}$. A multilayer perceptron $f_{\text{coop}}$ subsequently predicts cooperation probabilities: $\hat{c}_{t,i}=\text{softmax}(f_{\text{coop}}(\bar{H}_i))$. The model is trained using cross-entropy loss over pedestrians and batches:
\begin{equation}
\small
\mathcal{L}_{\text{coop}} = -\frac{1}{B} \sum_{b=1}^{B} \sum_{i=1}^{M} \sum_{j \in \{0,1\}} y_{b,i,j} \log \hat{c}_{b,i,j}
\label{eq:coop_loss}
\end{equation}
where $B$ is the batch size and $y_{b,i,j}$ denotes the ground-truth label. Cooperation labels are automatically generated in simulation based on pedestrians’ reactive behaviors during interaction, without manual annotation.

\subsection{Reinforcement Learning for Dynamic Horizon Decision}
\subsubsection{Observation}
At time step \(t\), human--robot interactions in heterogeneous crowds are formalized as a spatio-temporal graph 
\(G_t = (V_t, E_t^S, E_t^T)\), comprising agent nodes \(V_t\), spatial edges \(E_t^S\), and temporal edges \(E_t^T\). 
The graph incorporates the predicted cooperation attributes \(c_{t,i}\) from Stage I, enabling the RL policy to reason about social heterogeneity.

\textbf{Agent Nodes $V_t$:} 
The node set is defined as 
\(V_t = \{s_t^r, s_{t,1}^p, ..., s_{t,n}^p\}\), where \(s_t^r\) represents the robot state and $s_{t,i=1:n}^p$ denotes the observable state of the $n$ pedestrians within range.

\begin{equation}
     s_t^r = (p_t^r, v_t^r, \theta_t, g, v_{\text{ref}}, r),
\label{eq:robot_obs}
\end{equation}  
where 
\(p_t^r = (p_{x,t}^r, p_{y,t}^r)\) is the robot position at time $t$,  
\(v_t^r = (v_{x,t}^r, v_{y,t}^r)\) is the robot velocity,  
\(\theta_t\) is the heading,  
\(g = (g_x, g_y)\) is the goal position,  
\(v_{\text{ref}}\) is the reference velocity, and  
\(r\) is the robot radius.

The pedestrian state $s_{t,i}^p$ is represented in the robot-relative coordinate frame and is augmented with the predicted cooperation attribute $c_{t,i}$ from Stage I to capture social heterogeneity:
\begin{equation}
    s_{t,i}^p = (\Delta p_{t,i}, \Delta v_{t,i}, c_{t,i}),
\label{eq:ped_obs}
\end{equation}  
where 
$\Delta p_{t,i} = (\Delta x_{t,i}, \Delta y_{t,i})$ is the position of pedestrian $i$ relative to the robot at time $t$,  
$\Delta v_{t,i} = (\Delta v_{x,t,i}, \Delta v_{y,t,i})$ is the velocity of pedestrian $i$ relative to the robot, and  
$c_{t,i} \in \{0,1\}$ indicates the predicted cooperation attribute from the spatio-temporal transformer.

\textbf{Spatial Edges $E_t^S$:}
Spatial edges encode interactions among visible agents, 
\(E_t^S = \{ e_{ij}^S \mid s_t^i, s_t^j \in V_t,\ i \neq j \}\), 
with \(e_{ij}^S\) capturing visibility-based adjacency. 
This design emphasizes salient local relations critical for navigation. 

\textbf{Temporal Edges $E_t^T$:}
Temporal edges \(E_t^T = \{ e_{i,t}^T \mid (s_{t-1}^i, s_t^i),\ s_t^i \in V_t \}\) 
connect the same agent across consecutive time steps, preserving motion continuity and encoding dynamic information such as velocity for long-horizon reasoning.

\subsubsection{Action}
The RL policy $\pi_\theta$ operates in a discrete action space and decides the MPC prediction horizon. 
At each time step, the action $h_t \in \mathcal{H} = \{1, 2, \dots, h_{\text{max}}\}$, where $h_{\text{max}}$ denotes the maximum allowable horizon.

\subsubsection{Reward}
The reward $r_t$ balances safety, efficiency, and social awareness. In addition to standard terminal, potential, and kinematic rewards, two task-specific signals are introduced: a visibility social reward incorporating predicted cooperation from Stage I, and a prediction horizon reward aligning the MPC horizon decision in Stage II.

\begin{equation}
r_t = r_t^\text{term} + r_t^\text{pot} + r_t^\text{kin} + r_t^\text{horizon} + r_t^\text{vis\_social}.
\label{eq:total_reward}
\end{equation}

\textbf{Prediction Horizon Reward:}
To align with Stage II, the policy is encouraged to decide an appropriate MPC prediction horizon $h_t$:
\begin{equation}
r_t^\text{horizon} = - \lambda_h (h_\text{max} - h_t),
\label{eq:horizon_reward}
\end{equation}
where $\lambda_h$ is the weight coefficient for the horizon reward. This term is balanced by the visibility social reward, which penalizes aggressive long-horizon decisions in the presence of non-cooperative pedestrians.
The horizon decision balances efficiency with social compliance: longer horizons provide greater foresight, while the visibility social reward constrains aggressive actions when non-cooperative pedestrians are present.

\textbf{Visibility Social Reward:}  
The robot is rewarded for considering the cooperation of visible pedestrians:
\begin{equation}
r_t^\text{vis\_social} =
\begin{cases}
\lambda_\text{high} \cdot h_t^{\eta_\text{high}} \cdot \rho_\text{no-coop}, & \text{if } \rho_\text{no-coop} > 0.5, \\
\lambda_\text{low} \cdot h_t^{\eta_\text{low}} \cdot (1 - \rho_\text{no-coop}), & \text{otherwise},
\end{cases}
\label{eq:social_reward}
\end{equation}
where $\rho_\text{no-coop}$ is the fraction of non-cooperative pedestrians within the robot's sensor view. The coefficients $\lambda_\text{high}, \lambda_\text{low}$ are scaling factors, and the exponents $\eta_\text{high}, \eta_\text{low}$ modulate the sensitivity of the reward to the MPC horizon.

\textbf{Terminal Reward:}  
Collisions, timeout, and goal achievement are unified into a single terminal reward:
\begin{equation}
r_t^\text{term} =
\begin{cases}
+10, & \text{goal reached}, \\
-20, & \text{collision or timeout}, \\
0, & \text{otherwise}.
\end{cases}
\label{eq:terminal_reward}
\end{equation}

\textbf{Potential Reward:}  
To incentivize efficient progress toward the goal, a potential-based reward is defined:
\begin{equation}
r_t^\text{pot} = -\lambda_\text{pot} \left( \Phi_t - \Phi_{t-1} \right),
\label{eq:potential_reward}
\end{equation}
where $\Phi_t = \| p_t^r - g \|$ denotes the robot's distance to the goal at time $t$, and $\lambda_\text{pot}$ is a constant scaling factor.

\textbf{Kinematic Reward:}  
To discourage aggressive or backward motions, a kinematic penalty is applied:
\begin{equation}
r_t^\text{kin} = - \lambda_r w_t^2 - \lambda_v \max(0, -v_t),
\label{eq:kinematic_reward}
\end{equation}
where $w_t$ and $v_t$ denote the rotational and translational control commands, and $\lambda_r, \lambda_v$ are positive weight coefficients.

\subsubsection{Algorithm} 
The method employs Proximal Policy Optimization (PPO) \cite{PPO} for training. The policy determines the prediction horizon for MPC optimization, thereby better accommodating varying environments.

\subsection{Optimal-Horizon MPC with Social Awareness}
This stage integrates predicted cooperation attributes $c_{t,i}$ and the optimal horizon $h_t$ into a socially-aware MPC framework. By adapting safety distances and cost penalties, the robot maintains social compliance while navigating among heterogeneous pedestrians.

\subsubsection{MPC Formulation}
The MPC problem is formulated to account for pedestrian cooperation uncertainty:
\label{eq:mpc_social_aware_twocolumn}
\begin{align}
u^*_{t:t+h_t-1|t} 
&= \arg\min_{\mathbf{u}_{t:t+h_t-1|t}} 
\Biggl\{ 
    \mathbb{E}_{c_{t,i}} \Bigl[ 
        p(\mathbf{x}_{t+h_t|t}, c_{t,i}) \notag \\
&\qquad + \sum_{k=0}^{h_t-1} q(\mathbf{x}_{t+k|t}, u_{t+k|t}, c_{t,i}) 
    \Bigr] 
\Biggr\} \\
\text{s.t.} \quad
& \mathbf{x}_{t+k+1|t} = f(\mathbf{x}_{t+k|t}, u_{t+k|t}), \quad k=0,\dots,h_t-1, \notag\\
& \mathbf{x}_{t+k|t} \in \mathcal{X}, \quad 
  u_{t+k|t} \in \mathcal{U}, \quad k=0,\dots,h_t-1, \notag\\
& \mathbf{x}_{t|t} = \mathbf{x}_t, \quad 
  \mathbf{x}_{t+h_t|t} \in \mathcal{X}_f, \notag\\
& h_{\text{CBF}}^i(\mathbf{x}_{t+k|t}, c_{t,i}) \ge 0, 
  \quad \forall i, k. \notag
\end{align}
where $p(\cdot)$, $q(\cdot)$, and $f(\cdot)$ denote terminal cost, stage cost, and robot dynamics, respectively. The expectation $\mathbb{E}_{c_{t,i}}[\cdot]$ is taken over the cooperation probabilities $P(c_{t,i})$ predicted in Stage I.

\subsubsection{Cost Functions}
The stage cost balances goal-reaching, control effort, and social interaction:
\begin{equation}
\begin{split}
q(\mathbf{x}_{t+k|t}, u_{t+k|t}, c_{t,i}) 
&= (\mathbf{x}_{t+k|t}-\mathbf{x}_{\text{goal}})^\top Q (\mathbf{x}_{t+k|t}-\mathbf{x}_{\text{goal}}) \\
&\quad + u_{t+k|t}^\top R u_{t+k|t} \\
&\quad + \sum_i \phi(\mathbf{x}_{t+k|t}, p_i, c_{t,i}),
\end{split}
\label{eq:stage_cost}
\end{equation}
where $Q$ and $R$ are weighting matrices. The cooperation-aware social cost $\phi(\cdot)$ is defined as:
\begin{equation}
\phi(\mathbf{x}, p_i, c_{t,i}) = \eta_i \cdot \exp\left( -\frac{\|p_r - p_i\|^2}{\sigma_{c_{t,i}}^2} \right)
\label{eq:social_cost_function}
\end{equation}
where $\eta_i$ is a scaling factor and $\sigma_{c_{t,i}}$ modulates the influence range. A smaller $\sigma$ is assigned to cooperative pedestrians ($c_{t,i}=1$) to allow closer interaction, while a larger $\sigma$ for non-cooperative agents ensures conservative avoidance.

\subsubsection{Safety Constraints via Discrete-time Control Barrier Functions}
To ensure the forward invariance of the safe set $\mathcal{C}$, we employ the Discrete-time Control Barrier Function (DTCBF) framework. For each pedestrian $i$, we enforce the following linear constraint on the evolution of $h^i$:
\begin{equation}
h^i(\mathbf{x}_{t+k+1|t}, c_{t,i}) - h^i(\mathbf{x}_{t+k|t}, c_{t,i}) \ge -\gamma h^i(\mathbf{x}_{t+k|t}, c_{t,i})
\label{eq:dtcbf_condition}
\end{equation}
where $\gamma \in (0, 1]$ modulates the permissiveness of the safe set boundary. The barrier function $h^i$ is defined as:
\begin{equation}
h^i(\mathbf{x}_{t+k|t}, c_{t,i}) = \|p_r - p_i\| - \big( r_r + r_i + d_{\text{safety}}^i(c_{t,i}) \big)
\label{eq:cbf_definition}
\end{equation}
where $d_{\text{safety}}^i(c_{t,i}) = d_0 + c_{t,i}\, d_{\text{coop}} + (1-c_{t,i})\, d_{\text{non-coop}}$. Note that pedestrian positions $p_i$ over the horizon are linearly projected based on current velocities, while the margin $d_{\text{safety}}^i$ dynamically adapts to inferred social attributes. System stability is preserved through these DTCBF constraints, ensuring safe state transitions even under dynamic planning horizon adjustments.

\section{Experimental Results}
\subsection{Experimental Setup}
\subsubsection{Simulation Environment} 
We extend the CrowdNav \cite{CrowdNav} environment to a realistic 2D pedestrian simulation. The robot is equipped with a limited 5$m$ circular sensing range and a 360° field of view, instead of assuming full observability. To reflect real-world uncertainty, pedestrians may be partially occluded. Social diversity is introduced by modeling cooperative pedestrians whose trajectories adapt to the robot, and non-cooperative pedestrians whose behaviors remain unaffected and non-visible to the robot. All pedestrians follow the ORCA strategy under speed and curvature constraints. This setup requires the robot to reason about occlusions and heterogeneous social behaviors.
\subsubsection{Comparative and Ablation Experiments} 
We benchmark 4 classic methods: reaction-based methods ORCA and SF, the trajectory-based method GST-RL, and the learning-based methods DS-RNN.

To assess the contribution of each component, two ablation experiments are performed:  
(i) Removing the RL-based horizon decision by adopting a fixed prediction horizon while keeping other modules unchanged;  
(ii) Removing the cooperation prediction module, which excludes pedestrian behavior classification, similar to prior methods.

\subsubsection{Evaluation Metrics} 
Performance is assessed across scenarios with low, medium, and high interaction levels, characterized by varying ratios of cooperative and non-cooperative pedestrians. Each model is tested on 250 random unseen cases and evaluated based on success rate (SR\%), collision rate (CR\%), timeout rate (OR\%), average navigation time (ANT$s$), average trajectory length (ATL$m$), and average intrusion ratio (AIR\%). AIR measures the percentage of timesteps during which the robot comes closer than the safety distance to any pedestrian. A lower AIR indicates that the robot maintains better social comfort distances with pedestrians and achieves more natural social navigation behavior.

\subsection{Comparative Experiments}
\begin{table*}
  \centering
   \caption{Comparison of state-of-the-art methods in multiple navigation scenarios with varying degrees of interactivity.}
  \label{tab:baseline_evaluation}
  \scalebox{1.0}{
  
  \renewcommand{\arraystretch}{1.8}
  \begin{threeparttable}
  \begin{tabular}{l||ccccc||ccccc||ccccc}
  \toprule
\multirow{2}{*}{\textbf{Method}}  & \multicolumn{5}{c||}{\textbf{Low Interaction Scenarios}} & \multicolumn{5}{c||}{\textbf{Mid Interaction Scenarios}} & \multicolumn{5}{c}{\textbf{High Interaction Scenarios}} \\
\cline{2-16}
&\textbf{SR$\uparrow$} &\textbf{CR$\downarrow$} &\textbf{OR$\downarrow$} &\textbf{ANT$\downarrow$} &\textbf{AIR$\downarrow$} &\textbf{SR$\uparrow$} &\textbf{CR$\downarrow$} &\textbf{OR$\downarrow$} &\textbf{ANT$\downarrow$} &\textbf{AIR$\downarrow$} &\textbf{SR$\uparrow$} &\textbf{CR$\downarrow$} &\textbf{OR$\downarrow$} &\textbf{ANT$\downarrow$} &\textbf{AIR$\downarrow$}
\\
\hline
\hline
ORCA\cite{ORCA1} &66.4 & 24.4 & 9.2 &\textbf{12.18} &28.19 &75.6 &15.2 &9.2 &\textbf{12.66} &25.75 &84.8 &7.2 &8.0 &\textbf{12.21} &24.24\\
\hline
SF\cite{SF} &18.0 &72.4 &9.6 &14.69 &26.80 &30.4 &56.4 &13.2 &14.16 &24.23 &42.0 &45.2 &12.8 &14.64 &22.51\\
\hline
\hline
DS-RNN\cite{DS-RNN} &51.2 &42.4 &\textbf{6.4} &12.60 &\textbf{21.54} &56.4 &36.8 &\textbf{6.8} &13.06 &\textbf{20.97} &70.0 &21.2 &8.8 &12.42 &\textbf{18.79}\\
\hline
GST-RL\cite{GST-RL} &\textbf{82.8} &\cellcolor{green!8}10.0 &7.2 &13.78 &\cellcolor{green!8}12.54 &\textbf{84.8} &\cellcolor{green!8}7.2 &8.0 &13.30 &\cellcolor{green!8}12.09 &\textbf{89.6} &\textbf{5.6} &\textbf{4.8} &12.42 &\cellcolor{green!8}11.51\\
\hline
\hline

Ours & \cellcolor{green!8}87.6 &\textbf{11.6} &\cellcolor{green!8}0.8 &\cellcolor{green!8}10.99 &25.84 &\cellcolor{green!8}92.0 &\textbf{7.2} &\cellcolor{green!8}0.8 &\cellcolor{green!8}10.33 &23.25 &\cellcolor{green!8}96.4 &\cellcolor{green!8}2.8 &\cellcolor{green!8}0.8 &\cellcolor{green!8}9.88 &20.40 \\
\bottomrule
\end{tabular}

\begin{tablenotes}
\footnotesize
    \item \textit{- \textsuperscript{1}Low Interaction Scenarios: 0 coop. peds. and 20 non-coop. peds. \textsuperscript{2}Mid Interaction Scenarios: 5 coop. peds. and 15 non-coop. peds. \textsuperscript{3}High Interaction Scenarios: 10 coop. peds. and 10 non-coop. peds.}
    \item \textit{-The best performance for each metric is highlighted in green, while the second-best performance is shown in bold.}
\end{tablenotes}
\end{threeparttable}
}
\end{table*}

\subsubsection{Quantitative Evaluation}
We evaluated the baseline methods and our proposed approach across multiple interaction scenarios based on the aforementioned metrics, as shown in Table \ref{tab:baseline_evaluation}.

ORCA demonstrates strong performance with SR of 66.4\% to 84.8\% and the shortest ANT, but suffers from a higher CR of 7.2\% to 24.4\% and AIR of 24.24\% to 28.19\%. Note that ORCA falls short of a 100\% success rate as the reciprocity assumption is often violated by non-cooperative pedestrians, leading to local minima or frozen states in dense crowds~\cite{GST-RL}. SF performs poorly across all scenarios with a low SR of 18.0\% to 42.0\% and a high CR of 45.2\% to 72.4\%.

Among learning-based methods, GST-RL achieves the best performance with SR of 82.8\% to 89.6\% and low CR of 5.6\% to 10.0\%, while DS-RNN shows moderate performance with good OR but limited SR.

Our approach outperforms all baselines, achieving the highest SR of 87.6\% to 96.4\% and shortest ANT of 9.88s to 10.99s while maintaining low CR of 2.8\% to 11.6\%.

Compared to the best baseline method GST-RL, our approach improves SR by 6.8\% from 89.6\% to 96.4\%, reduces ANT by 2.33$s$ from 12.21s to 9.88s, and decreases OR by 4.0\% from 4.8\% to 0.8\%. These improvements validating that integrating latent social inference with optimal horizon control is essential for resolving social deadlocks in non-stationary crowd environments.

\subsubsection{Qualitative Experiments}
\begin{figure*}[t]
    \centering
    \begin{tabular}{c@{\hspace{-0.00\textwidth}}c@{\hspace{-0.00\textwidth}}c@{\hspace{-0.00\textwidth}}c@{\hspace{-0.00\textwidth}}c}
        \begin{overpic}[width=0.19\textwidth]{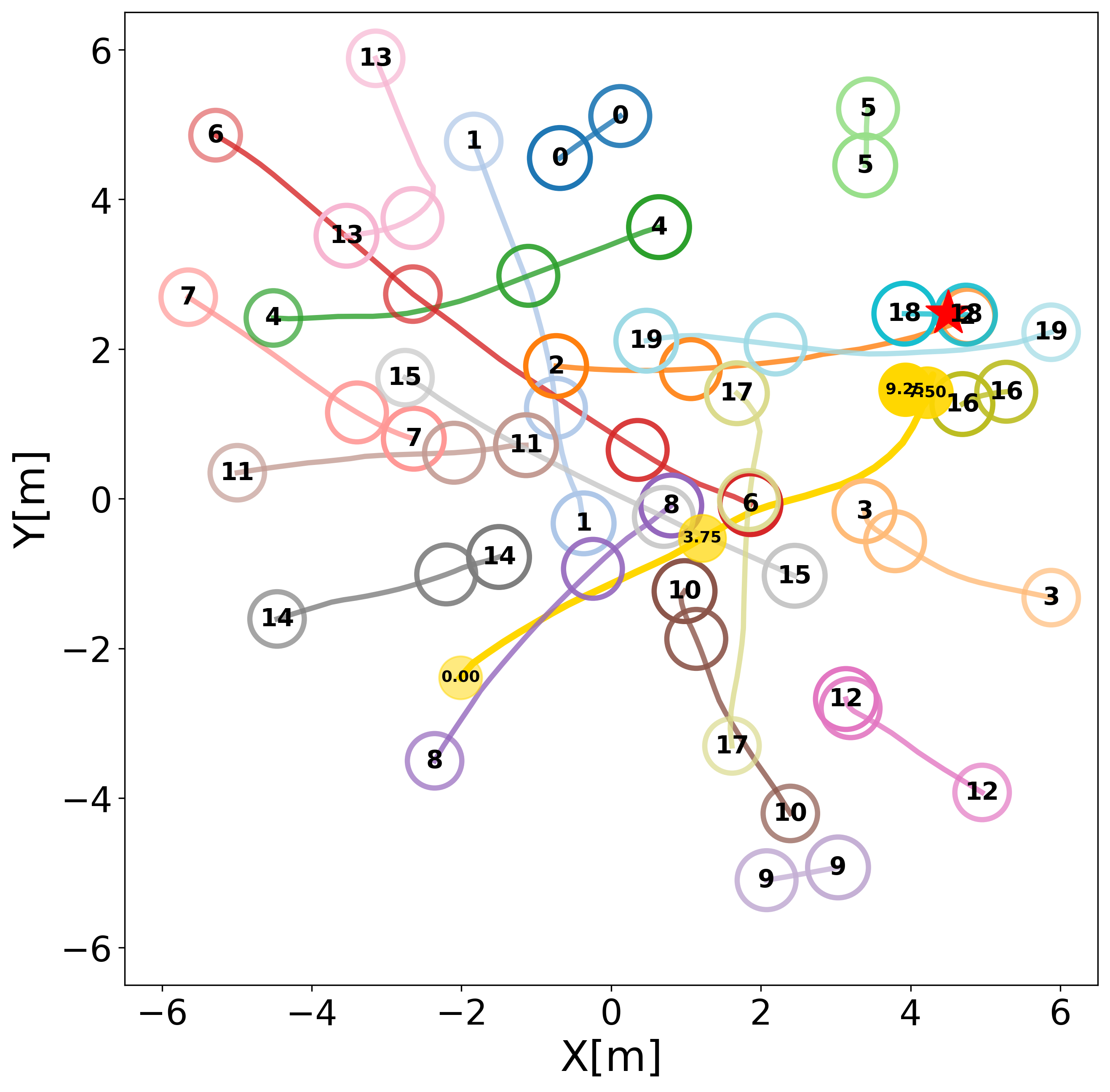}
            \put(62.5,88){\textcolor{red}{\textbf{\small Collision}}}
        \end{overpic} &
        \begin{overpic}[width=0.19\textwidth]{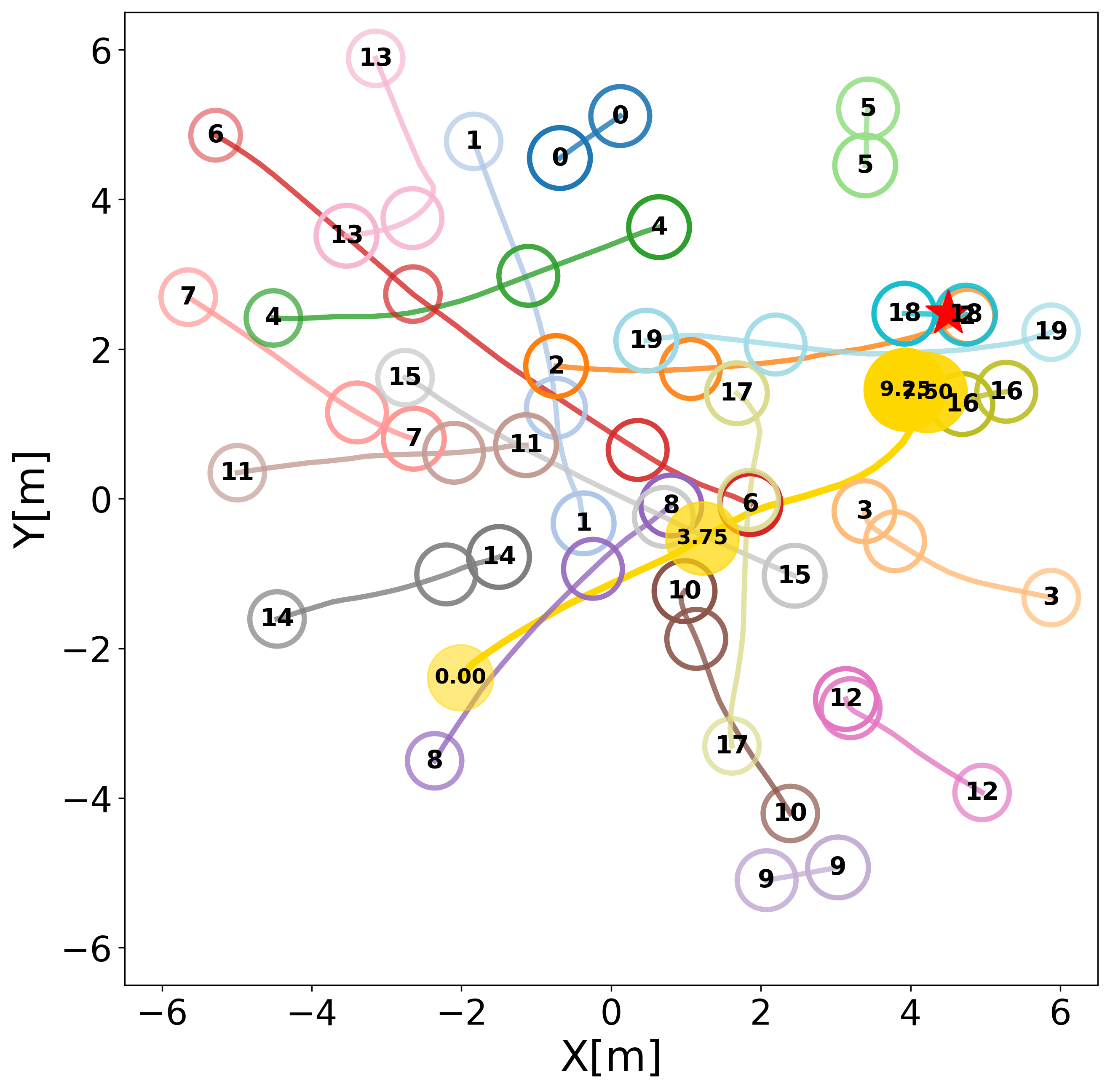}
            \put(62.5,88){\textcolor{red}{\textbf{\small Collision}}}
        \end{overpic} &
        \begin{overpic}[width=0.19\textwidth]{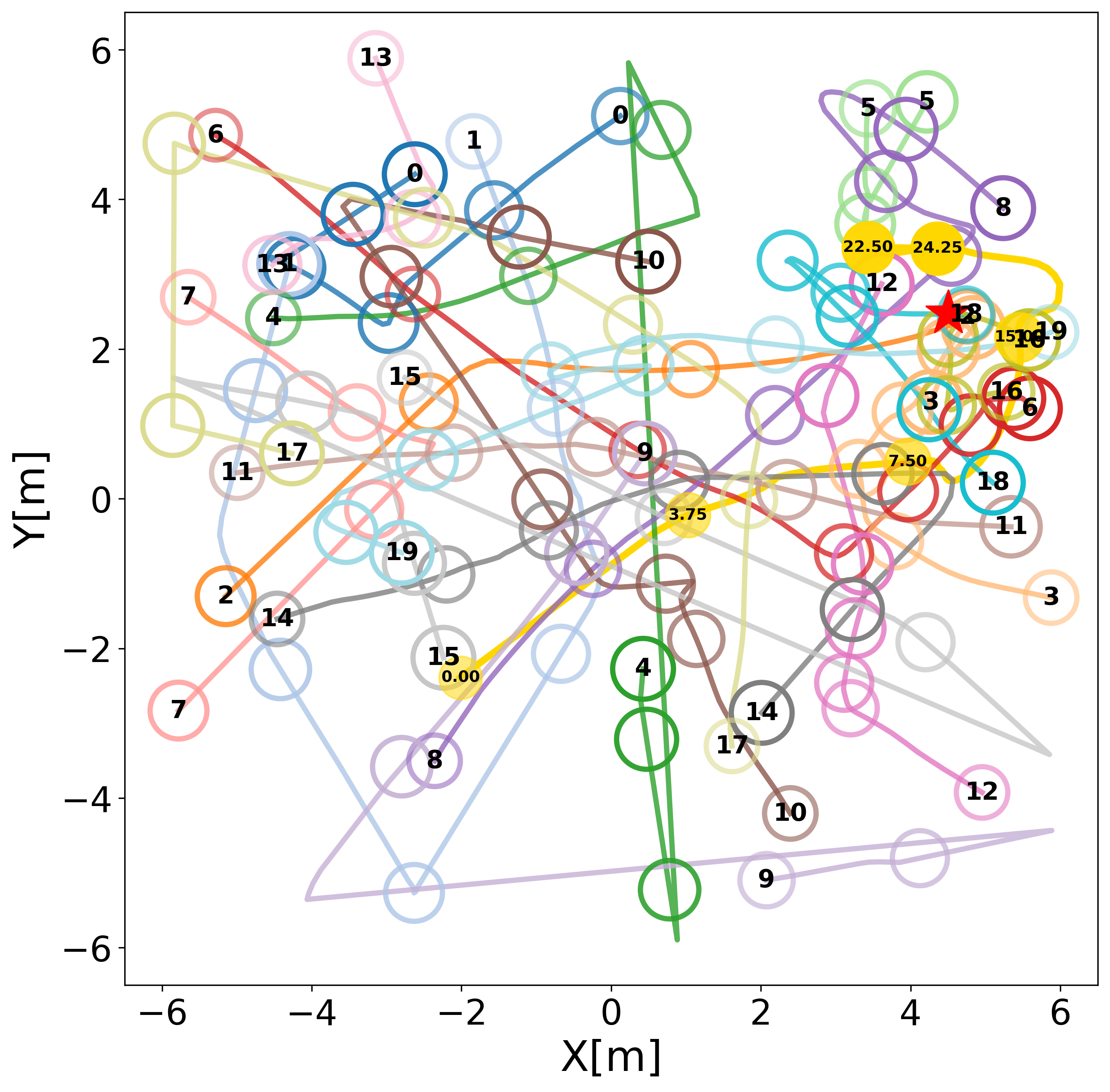}
            \put(64,88){\textcolor{orange}{\textbf{\small Timeout}}}
        \end{overpic} &
        \begin{overpic}[width=0.19\textwidth]{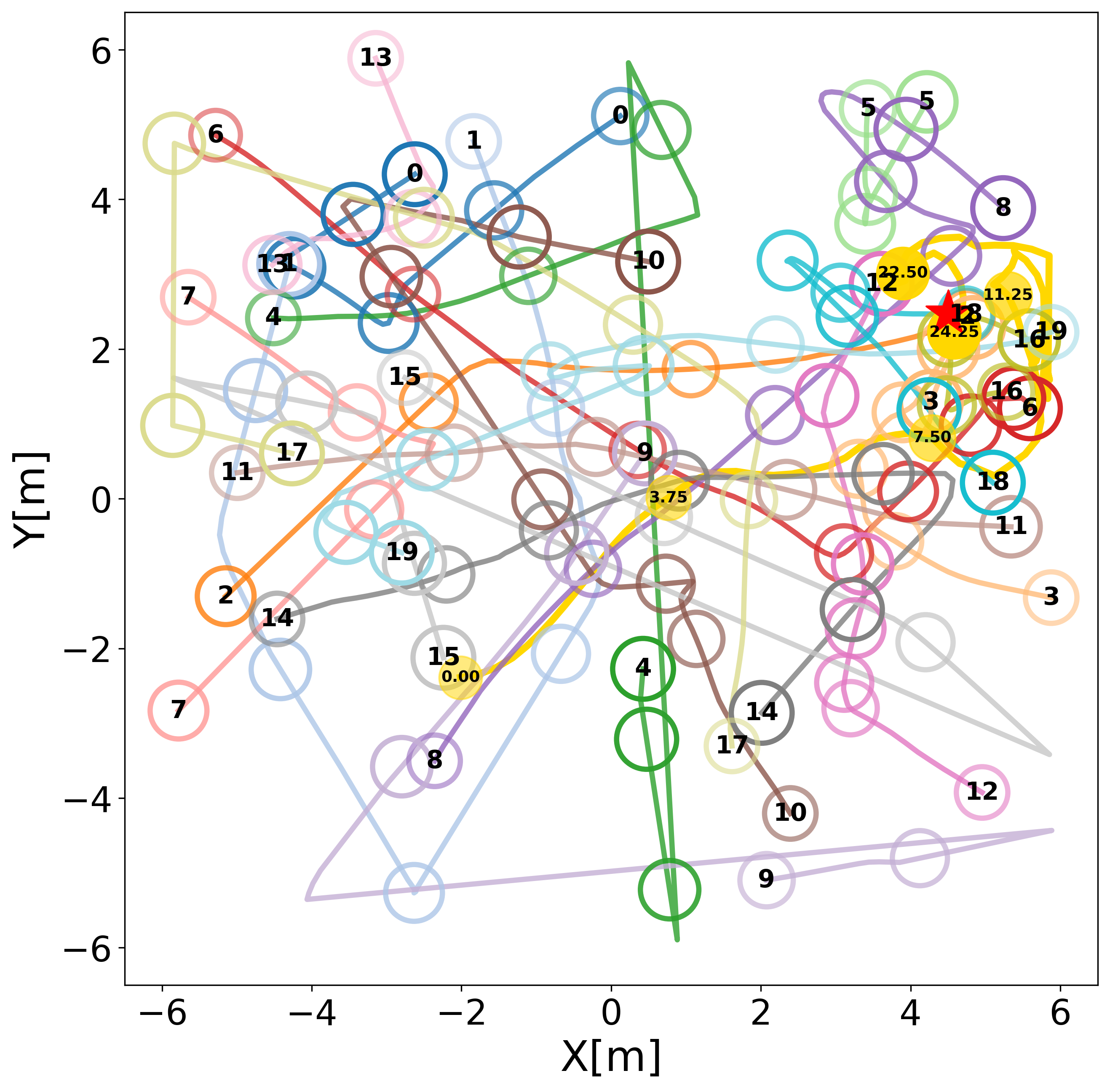}
            \put(64,88){\textcolor{orange}{\textbf{\small Timeout}}}
        \end{overpic} &
        \begin{overpic}[width=0.19\textwidth]{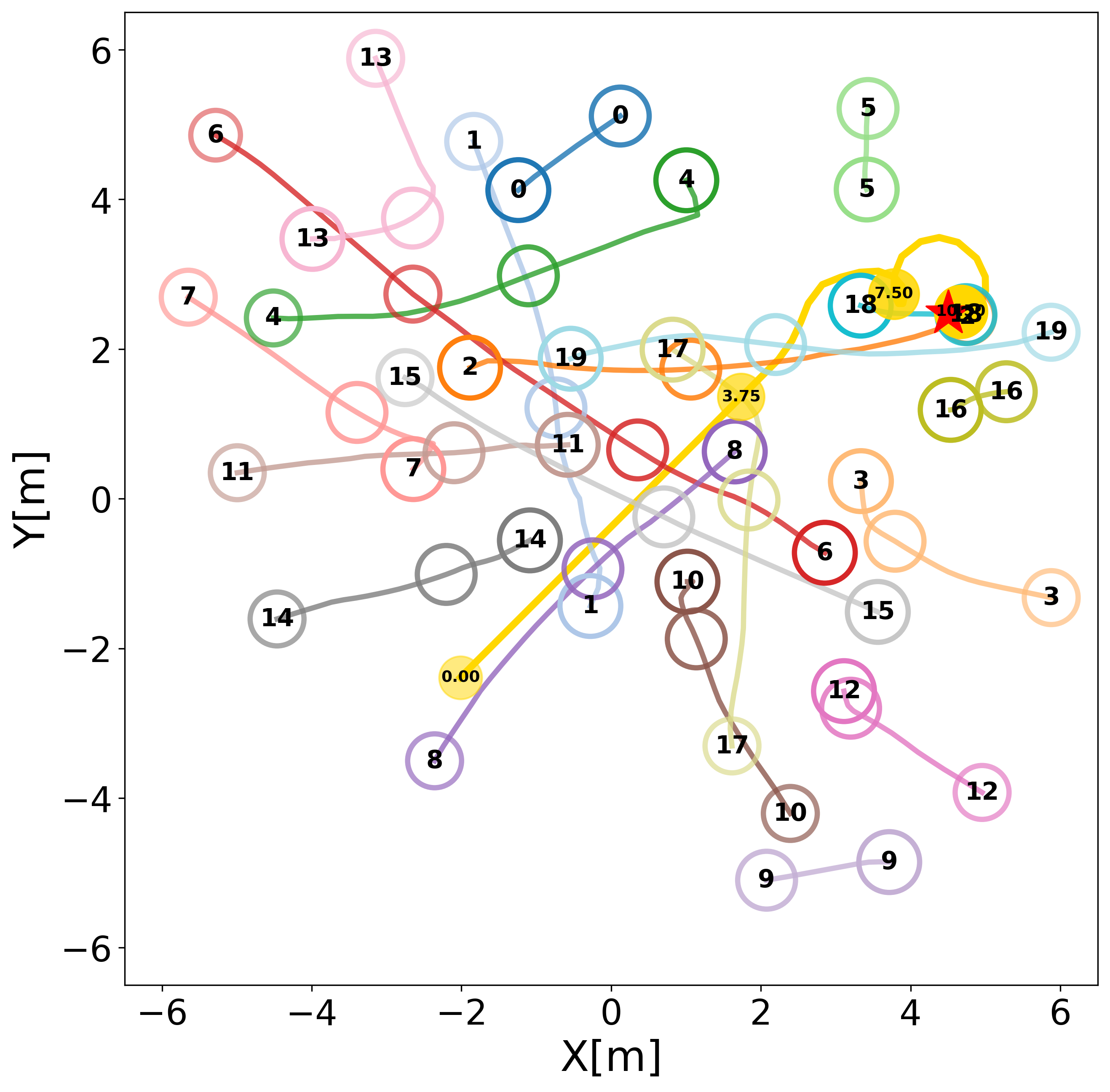}
            \put(67,88){\textcolor{darkgreen}{\textbf{\small Success}}}
        \end{overpic} \\
        (a) SF & (b) DS-RNN & (c) ORCA & (d) GST-RL & (e) Ours
    \end{tabular}
    \vspace{-0.1cm}
    \caption{Trajectories of all methods in low-interaction scenarios. All pedestrians in low-interaction scenarios are non-cooperative. The golden solid circles represent robot trajectories with timestamps. The hollow circles in different colors represent pedestrians, with pedestrian IDs shown at start and end points.}
    \label{fig:Low_Interaction_Comparison}
    \vspace{-0.4cm}
\end{figure*}

\begin{figure*}[t]
    \centering
    \begin{tabular}{@{}c@{\hspace{1.65em}}c@{\hspace{1.65em}}c@{\hspace{1.65em}}c@{\hspace{1.65em}}c@{\hspace{1.65em}}c@{\hspace{1.65em}}c@{}}
        \textcolor{red}{\textbf{\ \ \large$\star$}} \scriptsize Goal &
        \tikz[baseline=-0.5ex]{\draw[yellow,fill=yellow,line width=0.5pt] (0,0) circle (0.08cm);} \scriptsize Robot & 
        \tikz[baseline=-0.5ex]{\draw[darkgreen,fill=green!20,line width=0.5pt] (0,0) circle (0.08cm);} \scriptsize Cooperative pedestrian & 
        \tikz[baseline=-0.5ex]{\draw[red,fill=red!20,line width=0.5pt] (0,0) circle (0.08cm);} \scriptsize Non-cooperative pedestrian & 
        \tikz[baseline=-0.5ex]{\draw[gray,line width=0.8pt,dashed] (0,0) -- (0.6,0);} \scriptsize Radar rays & 
        \textcolor{yellow}{\textbf{---}} \scriptsize Robot trajectory & 
        \tikz[baseline=-0.5ex]{\draw[black,line width=0.8pt] (0,0) -- (0.4,0);} \scriptsize Human future trajectory  
    \end{tabular}
    \vspace{-8.0mm}  
\end{figure*}
\begin{figure}[t]
    \centering
    \setlength{\abovecaptionskip}{0.05cm}   

    \setlength{\fboxrule}{0.3pt}  
    \setlength{\fboxsep}{0.1mm}   

    \subfloat[ORCA]{ 
        \fbox{%
            \adjustbox{margin=-3.7mm -1.0mm -3.7mm 0.3mm}{ 
                \begin{tabular}{cccc}
                    \begin{overpic}[scale=0.089]{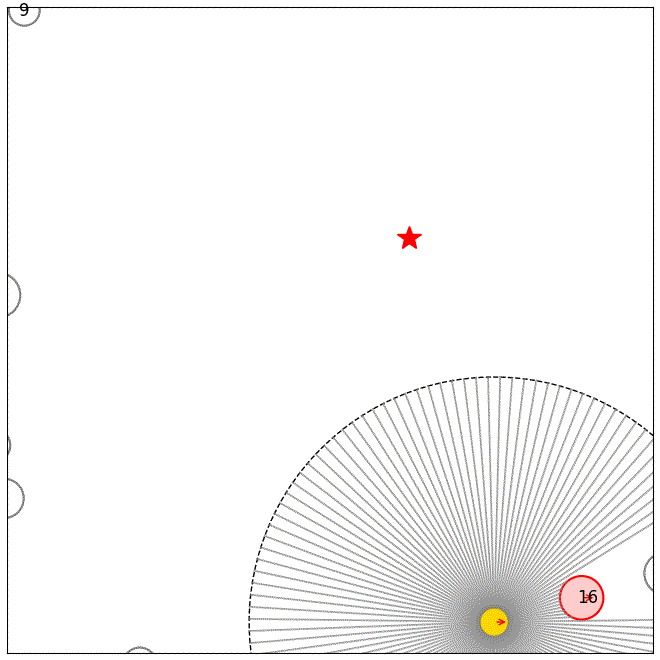}
                        \put(5,90){\textcolor{black}{\textbf{\tiny t=0s}}}
                    \end{overpic} &
                    \hspace{-1.5em} 
                    \begin{overpic}[scale=0.089]{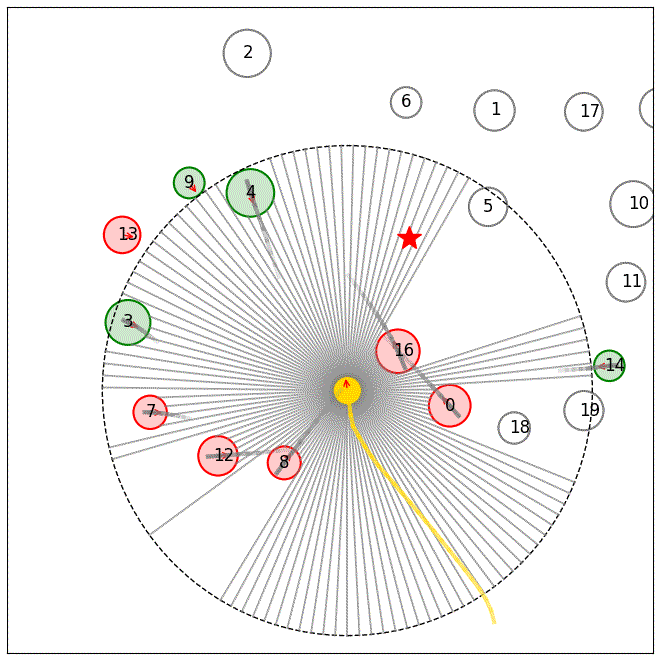}
                        \put(5,90){\textcolor{black}{\textbf{\tiny t=5.75s}}}
                    \end{overpic} &
                    \hspace{-1.5em} 
                    \begin{overpic}[scale=0.089]{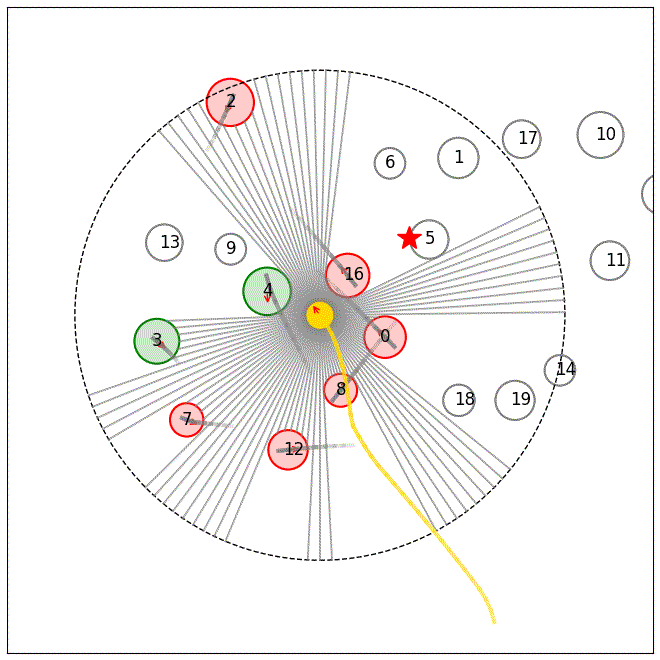}
                        \put(5,90){\textcolor{black}{\textbf{\tiny t=7.5s}}}
                    \end{overpic} &
                    \hspace{-1.5em} 
                    \begin{overpic}[scale=0.089]{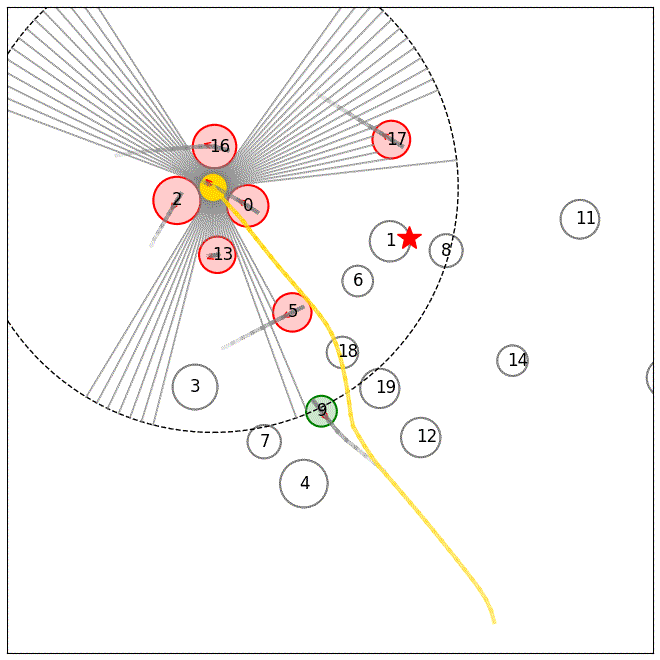}
                        \put(5,90){\textcolor{black}{\textbf{\tiny t=11.25s}}}
                        \put(65,90){\textcolor{red}{\textbf{\tiny Collision}}}  
                    \end{overpic} \\
                \end{tabular}
            }
        }
    } 
     \vspace{-3mm}  

    \subfloat[Ours]{
        \fbox{%
            \adjustbox{margin=-3.7mm -1.0mm -3.7mm 0.3mm}{ 
                \begin{tabular}{cccc}
                    \begin{overpic}[scale=0.089]{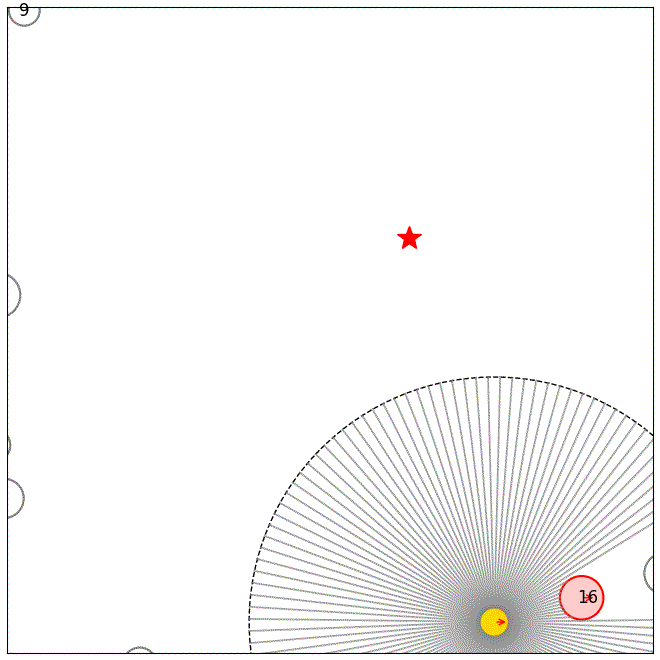}
                        \put(5,90){\textcolor{black}{\textbf{\tiny t=0s}}}
                    \end{overpic} &
                    \hspace{-1.5em} 
                    \begin{overpic}[scale=0.089]{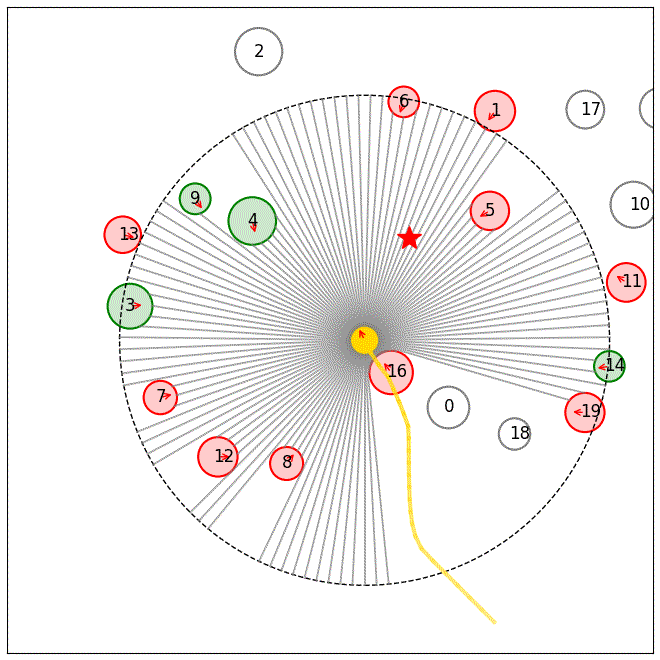}
                        \put(5,90){\textcolor{black}{\textbf{\tiny t=5.75s}}}
                    \end{overpic} &
                    \hspace{-1.5em} 
                    \begin{overpic}[scale=0.089]{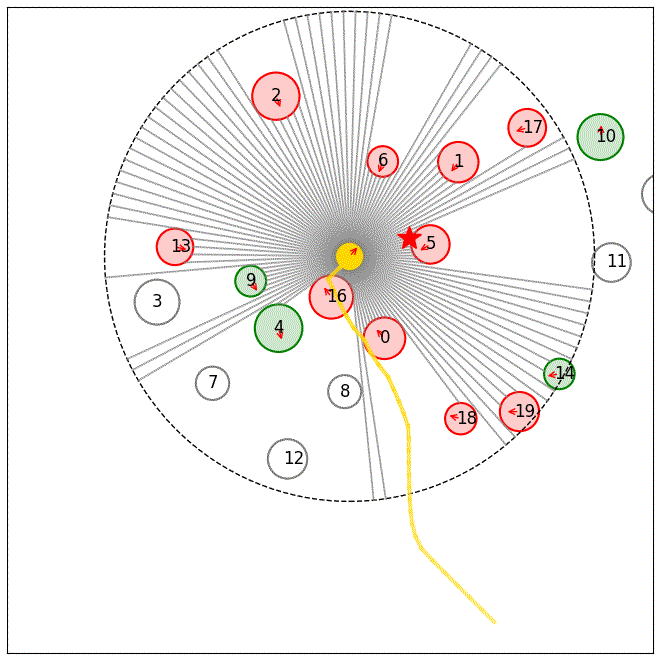}
                        \put(5,90){\textcolor{black}{\textbf{\tiny t=7.5s}}}
                    \end{overpic} &
                    \hspace{-1.5em} 
                    \begin{overpic}[scale=0.089]{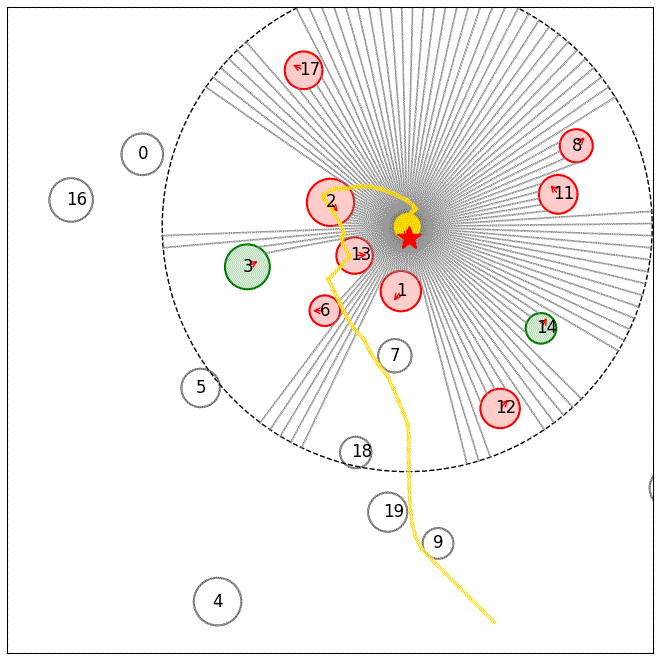}
                        \put(5,90){\textcolor{black}{\textbf{\tiny t=13.5s}}}
                        \put(70,90){\textcolor{darkgreen}{\textbf{\tiny Success}}}  
                    \end{overpic} \\
                \end{tabular}
            }
        }
    }

    \caption{Illustration of the navigation trajectory of ORCA and our proposed method in mid-interaction scenarios.}
    \label{fig:ORCA_Result}
\end{figure}
\begin{figure}[t]
    \setlength{\abovecaptionskip}{0.05cm}   

    \setlength{\fboxrule}{0.3pt}  
    \setlength{\fboxsep}{0.1mm}   

    \subfloat[GST-RL]{ 
        \fbox{%
            \adjustbox{margin=-3.7mm -1.0mm -3.7mm 0.3mm}{ 
                \begin{tabular}{cccc}
                    \begin{overpic}[scale=0.089]{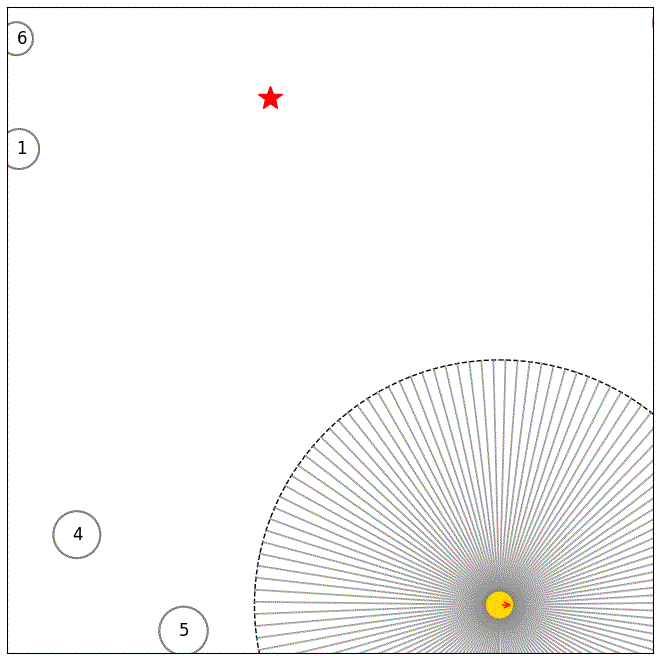}
                        \put(5,90){\textcolor{black}{\textbf{\tiny t=0s}}}
                    \end{overpic} &
                    \hspace{-1.5em} 
                    \begin{overpic}[scale=0.089]{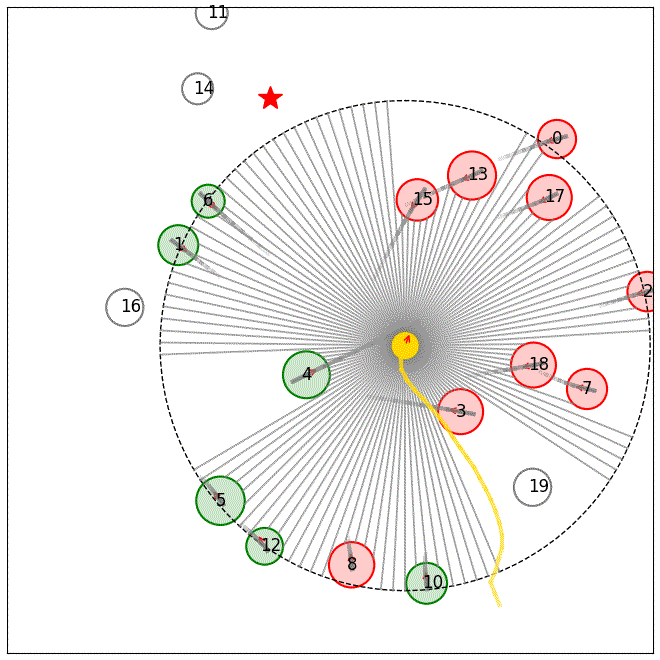}
                        \put(5,90){\textcolor{black}{\textbf{\tiny t=6s}}}
                    \end{overpic} &
                    \hspace{-1.5em} 
                    \begin{overpic}[scale=0.089]{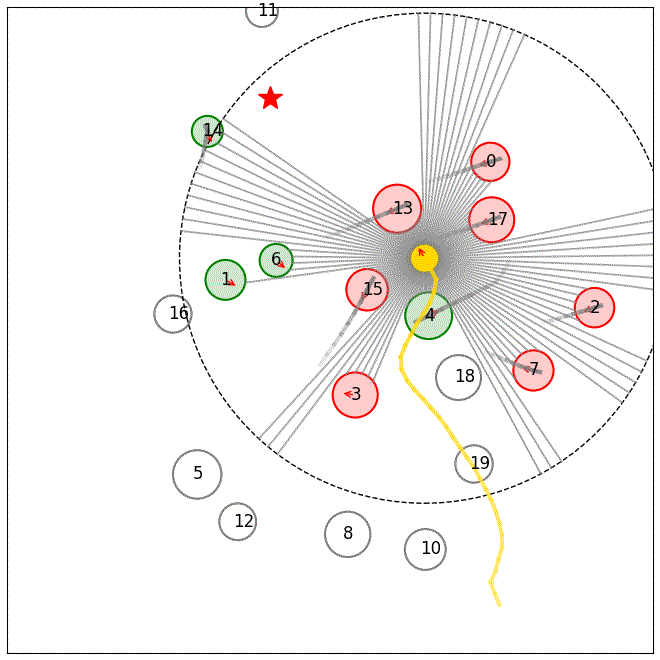}
                        \put(5,90){\textcolor{black}{\textbf{\tiny t=8s}}}
                    \end{overpic} &
                    \hspace{-1.5em} 
                    \begin{overpic}[scale=0.089]{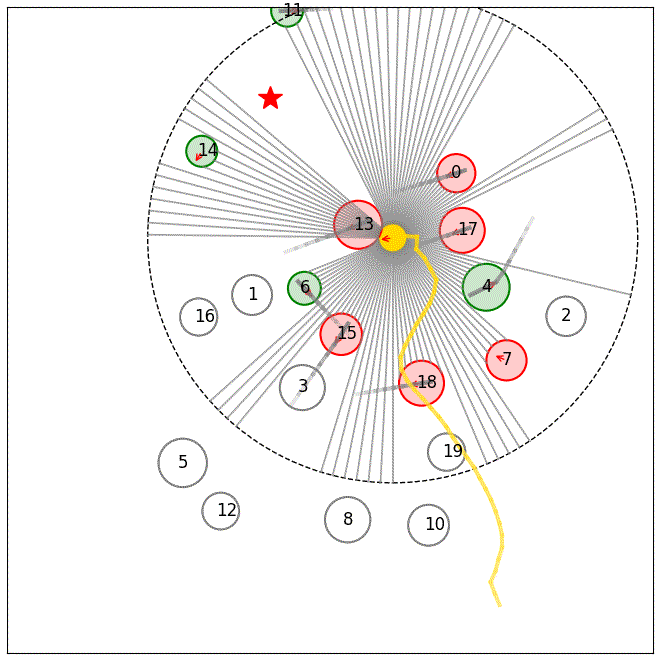}
                        \put(5,90){\textcolor{black}{\textbf{\tiny t=9s}}}
                        \put(65,90){\textcolor{red}{\textbf{\tiny Collision}}}  
                    \end{overpic} \\
                \end{tabular}
            }
        }
    } 
     \vspace{-3mm}  

    \subfloat[Ours]{
        \fbox{%
            \adjustbox{margin=-3.7mm -1.0mm -3.7mm 0.3mm}{ 
                \begin{tabular}{cccc}
                    \begin{overpic}[scale=0.089]{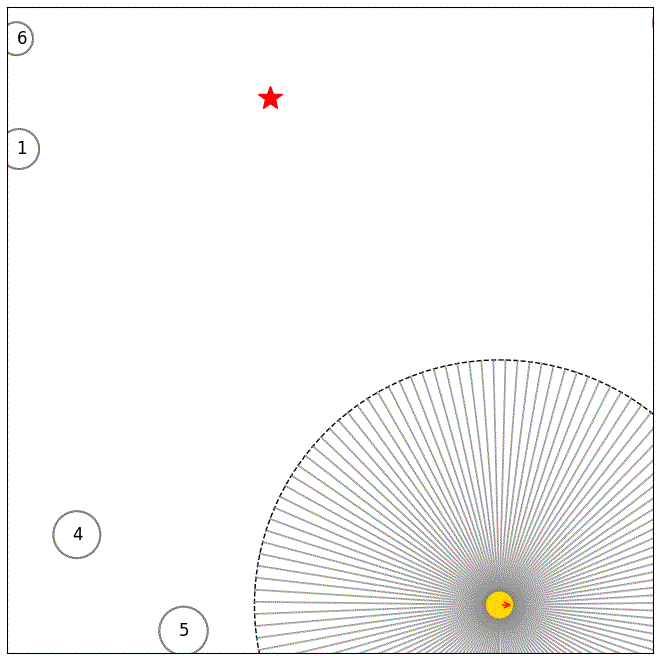}
                        \put(5,90){\textcolor{black}{\textbf{\tiny t=0s}}}
                    \end{overpic} &
                    \hspace{-1.5em} 
                    \begin{overpic}[scale=0.089]{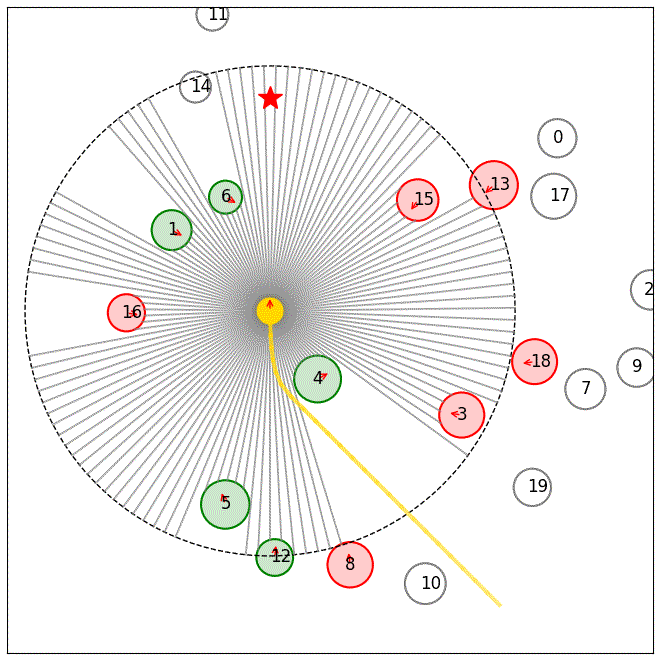}
                        \put(5,90){\textcolor{black}{\textbf{\tiny t=6s}}}
                    \end{overpic} &
                    \hspace{-1.5em} 
                    \begin{overpic}[scale=0.089]{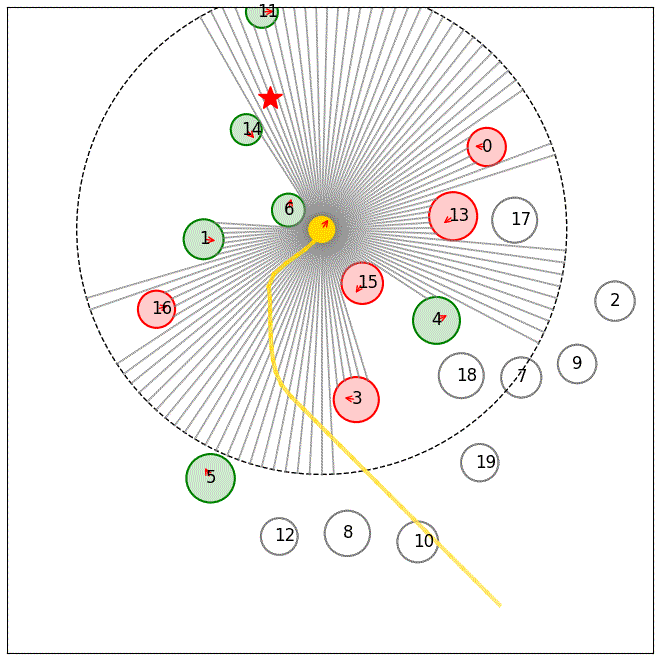}
                        \put(5,90){\textcolor{black}{\textbf{\tiny t=8s}}}
                    \end{overpic} &
                    \hspace{-1.5em} 
                    \begin{overpic}[scale=0.089]{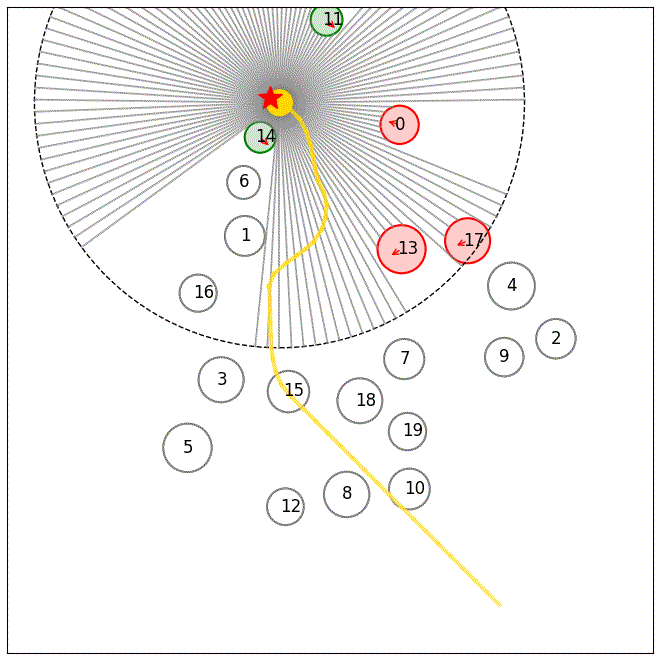}
                        \put(5,90){\textcolor{black}{\textbf{\tiny t=10.75s}}}
                        \put(70,90){\textcolor{darkgreen}{\textbf{\tiny Success}}}  
                    \end{overpic} \\
                \end{tabular}
            }
        }
    }

    \caption{Illustration of the navigation trajectory of GST-RL and our proposed method in high-interaction scenarios.}
    \label{fig:GST_RL_Result}
\end{figure}
The superiority of our approach is further demonstrated through qualitative experiments. In low-interaction scenarios, where the number of non-cooperative pedestrians increases and the drivable area decreases, SF \cite{SF} performs poorly due to lack of effective obstacle avoidance strategies and frequently experiences collisions in complex environments. DS-RNN \cite{DS-RNN}, despite being learning-based, attempts to avoid pedestrians but struggles with complex crowd dynamics and frequently experiences collisions due to inadequate handling of non-cooperative interactions. ORCA \cite{ORCA1} tends to make small movements, often leading to timeouts. GST-RL \cite{GST-RL} predicts pedestrian trajectories to avoid collisions, but becomes overly cautious and frequently experiences timeouts in challenging scenarios. In contrast, our method dynamically adjusts the prediction horizon based on crowd density and complexity, enabling optimal navigation strategies for different pedestrian densities and safely reaching the goal as shown in Fig. \ref{fig:Low_Interaction_Comparison}.

In mid-interaction scenarios, where some pedestrians become cooperative, ORCA \cite{ORCA1} demonstrates improved performance compared to low-interaction scenarios, but reacts too late to dynamic scenarios, leading to entrapment within crowds and subsequent collisions. Our method accelerates through dense crowds by circumventing non-cooperative pedestrians, thereby reaching the goal and mitigating the Frozen Robot Problem, as shown in Fig. \ref{fig:ORCA_Result}.

In high-interaction scenarios, GST-RL \cite{GST-RL} still treats cooperative pedestrians as dynamic obstacles. To avoid them, its trajectory becomes distorted and falls into local minima, leading to collisions. Our method coordinates with cooperative pedestrians, avoiding unnecessary yielding and searching for the fastest and smoothest collision-free trajectory, as shown in Fig. \ref{fig:GST_RL_Result}.

\section{Conclusions}
In this paper, we proposed a socially compliant robot navigation framework for dynamic and heterogeneous crowds, addressing the Frozen Robot Problem in partially observable environments. By inferring pedestrian cooperation attributes via a spatio-temporal Transformer and treating the MPC prediction horizon as a socially conditioned decision variable optimized through reinforcement learning, the proposed method adapts planning foresight to varying crowd density and interaction complexity in real time. Extensive experiments demonstrate that optimal horizon control significantly improves navigation performance, achieving higher success rates, lower collision rates, and shorter navigation time compared to state-of-the-art methods, while maintaining safety and social compliance in crowded environments.

Additional ablation studies and extended analyses will be provided in a subsequent version.

\bibliographystyle{IEEEtran}
\bibliography{references.bib}

@INPROCEEDINGS{everett2018motion,
  author={Everett, Michael and Chen, Yu Fan and How, Jonathan P.},
  booktitle={2018 IEEE/RSJ International Conference on Intelligent Robots and Systems (IROS)}, 
  title={Motion Planning Among Dynamic, Decision-Making Agents with Deep Reinforcement Learning}, 
  year={2018},
  volume={},
  number={},
  pages={3052-3059},
  doi={10.1109/IROS.2018.8593871}}

@INPROCEEDINGS{chen2017socially,
  author={Chen, Yu Fan and Everett, Michael and Liu, Miao and How, Jonathan P.},
  booktitle={2017 IEEE/RSJ International Conference on Intelligent Robots and Systems (IROS)}, 
  title={Socially aware motion planning with deep reinforcement learning}, 
  year={2017},
  volume={},
  number={},
  pages={1343-1350},
  doi={10.1109/IROS.2017.8202312}}

@article{kretzschmar2016socially,
  title={Socially compliant mobile robot navigation via inverse reinforcement learning},
  author={ Kretzschmar, Henrik  and  Spies, Markus  and  Sprunk, Christoph  and  Burgard, Wolfram },
  journal={International Journal of Robotics Research},
  volume={35},
  number={11},
  pages={1289-1307},
  year={2016},
}

@ARTICLE{kong2025socially,
  author={Kong, Yuqi and Gong, Xiaofei and Wang, Yao and Yu, Jiajie and Lu, Bo and Chi, Wenzheng and Sun, Lining},
  journal={IEEE Transactions on Industrial Electronics}, 
  title={Socially Conscious Navigation of Mobile Robots Based on Deep Reinforcement Learning}, 
  year={2025},
  volume={72},
  number={12},
  pages={13542-13550},
  doi={10.1109/TIE.2025.3566742}}

@ARTICLE{kabzan2019learning,
  author={Kabzan, Juraj and Hewing, Lukas and Liniger, Alexander and Zeilinger, Melanie N.},
  journal={IEEE Robotics and Automation Letters}, 
  title={Learning-Based Model Predictive Control for Autonomous Racing}, 
  year={2019},
  volume={4},
  number={4},
  pages={3363-3370},
  doi={10.1109/LRA.2019.2926677}}

@ARTICLE{samavi2025sicnav,
  author={Samavi, Sepehr and Han, James R. and Shkurti, Florian and Schoellig, Angela P.},
  journal={IEEE Transactions on Robotics}, 
  title={SICNav: Safe and Interactive Crowd Navigation Using Model Predictive Control and Bilevel Optimization}, 
  year={2025},
  volume={41},
  number={},
  pages={801-818},
  doi={10.1109/TRO.2024.3484634}}

@INPROCEEDINGS{icra24_coop_mpc,
  author={Le, Viet-Anh and Tadiparthi, Vaishnav and Chalaki, Behdad and Mahjoub, Hossein Nourkhiz and D’Sa, Jovin and Moradi-Pari, Ehsan and Malikopoulos, Andreas A.},
  booktitle={2024 IEEE International Conference on Robotics and Automation (ICRA)}, 
  title={Multi-Robot Cooperative Navigation in Crowds: A Game-Theoretic Learning-Based Model Predictive Control Approach}, 
  year={2024},
  volume={},
  number={},
  pages={4834-4840},
  doi={10.1109/ICRA57147.2024.10611204}}

@INPROCEEDINGS{icra24_drcc_mpc,
  author={Ryu, Kanghyun and Mehr, Negar},
  booktitle={2024 IEEE International Conference on Robotics and Automation (ICRA)}, 
  title={Integrating Predictive Motion Uncertainties with Distributionally Robust Risk-Aware Control for Safe Robot Navigation in Crowds}, 
  year={2024},
  volume={},
  number={},
  pages={2410-2417},
  doi={10.1109/ICRA57147.2024.10610404}}

@article{shamsah2024socially,
  title={Socially Acceptable Bipedal Robot Navigation via Social Zonotope Network Model Predictive Control},
  author={Abdulaziz Shamsah and Krishanu Agarwal and Nigam Katta and Abirath Raju and Shreyas Kousik and Ye Zhao},
  journal={IEEE Transactions on Automation Science and Engineering},
  year={2024},
  volume={22},
  pages={10130-10148},
  url={https://api.semanticscholar.org/CorpusID:270711502}
}

@ARTICLE{FRP,
  author={Sathyamoorthy, Adarsh Jagan and Patel, Utsav and Guan, Tianrui and Manocha, Dinesh},
  journal={IEEE Robotics and Automation Letters}, 
  title={Frozone: Freezing-Free, Pedestrian-Friendly Navigation in Human Crowds}, 
  year={2020},
  volume={5},
  number={3},
  pages={4352-4359},
  doi={10.1109/LRA.2020.2996593}}

@INPROCEEDINGS{chen2020relational,
  author={Chen, Changan and Hu, Sha and Nikdel, Payam and Mori, Greg and Savva, Manolis},
  booktitle={2020 IEEE/RSJ International Conference on Intelligent Robots and Systems (IROS)}, 
  title={Relational Graph Learning for Crowd Navigation}, 
  year={2020},
  volume={},
  number={},
  pages={10007-10013},
  doi={10.1109/IROS45743.2020.9340705}}

@ARTICLE{mestres2024distributed,
  author={Mestres, Pol and Nieto-Granda, Carlos and Cortés, Jorge},
  journal={IEEE Robotics and Automation Letters}, 
  title={Distributed Safe Navigation of Multi-Agent Systems Using Control Barrier Function-Based Controllers}, 
  year={2024},
  volume={9},
  number={7},
  pages={6760-6767},
  doi={10.1109/LRA.2024.3414268}}

@ARTICLE{tan2022distributed,
  author={Tan, Xiao and Dimarogonas, Dimos V.},
  journal={IEEE Control Systems Letters}, 
  title={Distributed Implementation of Control Barrier Functions for Multi-agent Systems}, 
  year={2022},
  volume={6},
  number={},
  pages={1879-1884},
  doi={10.1109/LCSYS.2021.3133802}}

@ARTICLE{soman2025learning,
  author={Soman, Surya and Zanon, Mario and Bemporad, Alberto},
  journal={IEEE Transactions on Intelligent Transportation Systems}, 
  title={Learning-Based Stochastic Model Predictive Control for Autonomous Driving at Uncontrolled Intersections}, 
  year={2025},
  volume={26},
  number={2},
  pages={1538-1546},
  doi={10.1109/TITS.2024.3510041}}

@INPROCEEDINGS{le2024social,
  author={Le, Viet-Anh and Chalaki, Behdad and Tadiparthi, Vaishnav and Mahjoub, Hossein Nourkhiz and D’Sa, Jovin and Moradi-Pari, Ehsan},
  booktitle={2024 IEEE/RSJ International Conference on Intelligent Robots and Systems (IROS)}, 
  title={Social Navigation in Crowded Environments with Model Predictive Control and Deep Learning-Based Human Trajectory Prediction}, 
  year={2024},
  volume={},
  number={},
  pages={4793-4799},
  doi={10.1109/IROS58592.2024.10802371}}

@ARTICLE{liu2026height,
  author={Liu, Shuijing and Xia, Haochen and Pouria, Fatemeh Cheraghi and Hong, Kaiwen and Chakraborty, Neeloy and Hu, Zichao and Biswas, Joydeep and Driggs-Campbell, Katherine},
  journal={IEEE Transactions on Automation Science and Engineering}, 
  title={{HEIGHT}: Heterogeneous Interaction Graph Transformer for Robot Navigation in Crowded and Constrained Environments}, 
  year={2026},
  volume={23},
  pages={1211-1230},
  doi={10.1109/TASE.2025.3646588}
}

@ARTICLE{nguyen2024model,
  author={Nguyen, Minh-Nhat and Van, Mien and McIlvanna, Stephen and Sun, Yuzhu and Close, Jack and Olayemi, Kabirat and Jin, Yan},
  journal={IEEE Transactions on Intelligent Vehicles}, 
  title={Model-Free Safety Critical Model Predictive Control for Mobile Robot in Dynamic Environments}, 
  year={2024},
  volume={9},
  number={11},
  pages={6830-6842},
  doi={10.1109/TIV.2024.3389111}}

@ARTICLE{stefanini2024efficient,
  author={Stefanini, Elisa and Palmieri, Luigi and Rudenko, Andrey and Hielscher, Till and Linder, Timm and Pallottino, Lucia},
  journal={IEEE Robotics and Automation Letters}, 
  title={Efficient Context-Aware Model Predictive Control for Human-Aware Navigation}, 
  year={2024},
  volume={9},
  number={11},
  pages={9494-9501},
  doi={10.1109/LRA.2024.3461552}}

@ARTICLE{zhang2024heterogeneous,
  author={Zhang, Han and Zhang, Xiaohui and Feng, Zhao and Xiao, Xiaohui},
  journal={IEEE Robotics and Automation Letters}, 
  title={Heterogeneous Multi-Robot Cooperation With Asynchronous Multi-Agent Reinforcement Learning}, 
  year={2024},
  volume={9},
  number={1},
  pages={159-166},
  doi={10.1109/LRA.2023.3328448}}

@misc{PPO,
      title={Proximal Policy Optimization Algorithms}, 
      author={John Schulman and Filip Wolski and Prafulla Dhariwal and Alec Radford and Oleg Klimov},
      year={2017},
      eprint={1707.06347},
      archivePrefix={arXiv},
      primaryClass={cs.LG},
      url={https://arxiv.org/abs/1707.06347}, 
}

@INPROCEEDINGS{CrowdNav,
  author={Chen, Changan and Liu, Yuejiang and Kreiss, Sven and Alahi, Alexandre},
  booktitle={2019 International Conference on Robotics and Automation (ICRA)}, 
  title={Crowd-Robot Interaction: Crowd-Aware Robot Navigation With Attention-Based Deep Reinforcement Learning}, 
  year={2019},
  volume={},
  number={},
  pages={6015-6022},
  doi={10.1109/ICRA.2019.8794134}}

@inbook{ORCA1,
author = {van den Berg, Jur and Guy, Stephen and Lin, Ming and Manocha, Dinesh},
year = {2011},
month = {04},
pages = {3-19},
title = {Reciprocal n-Body Collision Avoidance},
volume = {70},
isbn = {978-3-642-19456-6},
journal = {Springer Tracts in Advanced Robotics},
doi = {10.1007/978-3-642-19457-3_1}
}

@INPROCEEDINGS{DS-RNN,
  author={Liu, Shuijing and Chang, Peixin and Liang, Weihang and Chakraborty, Neeloy and Driggs-Campbell, Katherine},
  booktitle={2021 IEEE International Conference on Robotics and Automation (ICRA)}, 
  title={Decentralized Structural-RNN for Robot Crowd Navigation with Deep Reinforcement Learning}, 
  year={2021},
  volume={},
  number={},
  pages={3517-3524},
  doi={10.1109/ICRA48506.2021.9561595}}

@inbook{SF,
author = {Helbing, Dirk and Molnar, Peter},
year = {1995},
month = {05},
pages = {},
title = {Social force model for pedestrian dynamics}
}

@INPROCEEDINGS{GST-RL,
  author={Liu, Shuijing and Chang, Peixin and Huang, Zhe and Chakraborty, Neeloy and Hong, Kaiwen and Liang, Weihang and McPherson, D. Livingston and Geng, Junyi and Driggs-Campbell, Katherine},
  booktitle={2023 IEEE International Conference on Robotics and Automation (ICRA)}, 
  title={Intention Aware Robot Crowd Navigation with Attention-Based Interaction Graph}, 
  year={2023},
  volume={},
  number={},
  pages={12015-12021},
  doi={10.1109/ICRA48891.2023.10160660}}

\end{document}